\documentclass[letterpaper, 10pt, conference]{ieeeconf}  

\IEEEoverridecommandlockouts    
\usepackage{ntheorem,amsmath,graphicx,float,caption}
\usepackage{subeqnarray,cases,subfigure,ulem,mathrsfs}

\usepackage{amssymb}
\usepackage{setspace}

\usepackage{relsize}
\usepackage[margin=0.75in]{geometry}

\title{\LARGE \bf Improving the Speed of Response of Learning Algorithms Using Multiple Models: An Introduction}
\author{{Kumpati S. Narendra, Snehasis Mukhopadyhay$^*$, and Yu Wang}\\
{Center for Systems Science, Yale University}
\thanks{*:Professor Snehasis Mukhopadhay is with the Department of Computer and Information Science, Indiana University - Purdue Unversity Indianapolis}}

\begin{document}
\maketitle
\section{Abstract}

This is the first of a series of papers that the authors propose to write on the subject of improving the speed of response of learning systems using multiple models. During the past two decades, the first author has worked on numerous methods for improving the stability, robustness, and performance of adaptive systems using multiple models and the other authors have  collaborated with him on some of them. Independently, they have also worked on several learning methods, and have considerable experience with their advantages and limitations. In particular, they are well aware that it is common knowledge that machine learning is in general very slow. Numerous attempts have been made by researchers to improve the speed of convergence of algorithms in different contexts. In view of the success of multiple model based methods in improving the speed of convergence in adaptive systems, the authors believe that the same approach will also prove fruitful in the domain of learning. In this paper, a first attempt is made to use multiple models for improving the speed of response of the simplest learning schemes that have been studied. i.e. Learning Automata.

\section{Introduction}
Learning is defined as any relatively permanent change in behavior resulting from past experience, and a learning system is characterized by its ability to improve in some sense, its behavior with time, tending towards an ultimate goal. In mathematical psychology, models of learning systems were developed about fifty years ago to explain behavior patterns among living organisms. These, in turn, were adapted to synthesize engineering systems, which could be considered to show "learning behavior". In fact, in 1971, Tsypkin [1] argued that seemingly diverse problems in pattern recognition, filtering, identification, and control could be treated in a unified manner as problems in learning, using probabilistic iterative methods.

Reinforcement learning aims to find the optimal decision (or decision rule) in uncertain environments, on the basis of qualitative and noisy on-line performance feedback provided by an environment, in the form of a reinforcement signal. During the past four decades, learning theory has grown into a huge field in which a very large number of models have been studied. While the models were initially static, the approaches developed were extended to Markov Decision Processes (MDP) with finite states, and later to stochastic nonlinear difference equation models for continuous state cases. Our objective in these reports is to successively consider the principal models suggested in the literature in all the different areas, and investigate how multiple model based methods can be developed to increase the speed and accuracy of learning in all of them.

\subsection{Problem Areas}
One of the earliest reinforcement learning models in the literature is the "learning automaton". An agent can perform one action out of a finite set of r actions at every instant into an environment. The environment responds to each action with either a "reward" or a "penalty". The probability of reward $d_i$ of each action $\alpha_i$(i=1,2,...r) is unknown. The objective is to device a procedure by which the agent learns the best action (i.e. $\alpha_{opt}$ corresponding to $d_{opt}=\max_{j} d_j$). A very large number of fixed structure and variable structure schemes have been reported in the literature and these have been treated in great detail in [2] by Narendra and Thathachar.
Developments in the basic learning automaton models in stationary environments led to non-stationary environments and eventually to Markov Decision Processes (MDP). 

In the latter, at every state in a finite Markov chain, an agent can use one of a finite number of actions. A transition matrix  defines the probabilities with which the state is transferred to any other state under a specific action, and corresponding to such a transfer, there is a reward attached to it. Both the transition matrix and reward probabilities are unknown. The objective is to determine the optimal action at every state, which results in the optimization of a performance criterion defined over a finite or infinite horizon. It is this problem which, in course of time, evolved to the optimal control problem of Markov processes.defined by unknown nonlinear difference equations. In such problems, identification of the system and optimization are carried out simultaneously at every stage. 

\underline{Model- Free (Direct) and Model-based (Indirect) Learning}
The large class of learning schemes which have evolved over the years can be broadly categorized into model-free or mode based classes. Model-free methods are direct methods where the optimal policy is learned without trying to identify the system. This is in sharp contrast to model based methods, where a model of the process is constantly updated and used to determine the optimal policy. This is entirely in the spirit of indirect adaptive control. We shall be interested in both direct and indirect schemes, while evaluating the effect of multiple models. In this report we shall consider only the simplest direct learning scheme i.e. the learning automaton. In Section II, a brief description of the scheme is provided and norms of behavior are described. In Section III, many of the important but simple learning schemes are reviewed and their properties are described.

Simulation results are provided to indicate the nature of convergence that can be expected in such schemes. Learning based on multiple models, which is the principal subject of this report demonstrates that significantly better responses can be achieved by using them.

\section{The Learning Automaton}
As stated in Section I, the learning automaton is the simplest scheme considered in detail in this report. It is a direct method and is concerned with an automaton choosing the best action in a random static environment, based on its responses. In spite of its simplicity, it  addresses many questions which are of fundamental importance in all the learning schemes treated in the reports follow.

We first consider the learning automaton in its simplest form. This is shown in Figure 1 where a random environment E and an automaton A are connected in a feedback loop. The random environment is described by a finite input set $\boldsymbol{\alpha}$=$\lbrace\alpha_1,\alpha_2,...\alpha_r \rbrace$ and an output set $\boldsymbol{\beta}$=$\lbrace 0,1\rbrace$. 1 is referred to as a "reward" and 0 as a "penalty". Corresponding to every action $\alpha_i$ is a reward probability $d_i$, where 
\begin{equation}
d_i=Pr[\beta=1|\alpha=\alpha_i],\;\;\;\;\;\alpha_i \in \boldsymbol{\alpha}
\end{equation}

The actions $\alpha_i$ can consequently be ordered using $d_i$ so that $\alpha_i$ is better than $\alpha_j$ if $d_i > d_j$, and the best action $\alpha_{opt}$ is the action that corresponds to $d_{opt}$, where $\max_id_i$=$d_{opt}$

\begin{figure}[H]
\includegraphics[width=0.5\textwidth,height=0.25\textwidth]{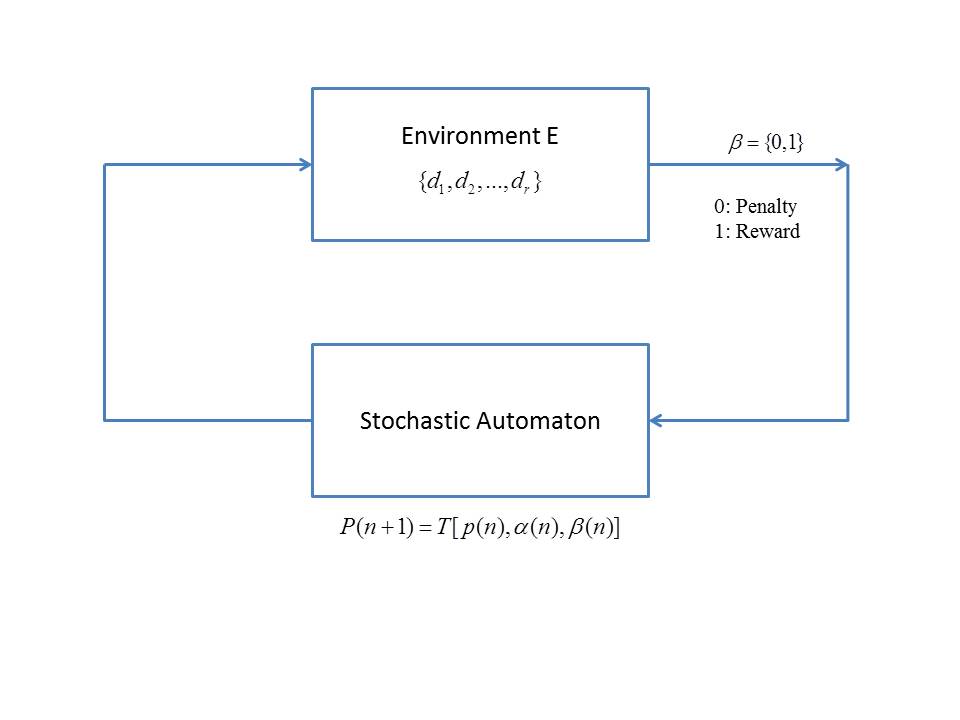}
\caption{The Learning Automaton}
\label{Fig1}
\end{figure}

\underline{The Automaton}: is a stochastic decision rule which, at every instant (n+1) selects an action from the set $\boldsymbol{\alpha}$, based on the response of the environment at step n, to an action $\alpha(n)$. In the first part of the twentieth century, deterministic automata were studied extensively in the (then) Soviet Union. However, our interest is strictly in stochastic automata. The operation of such an automaton can be described as follows.

At every instant 'n' the stochastic automaton chooses an action $\alpha(n)$, using a probability distribution on the finite action set. i.e.$\lbrace p_1(n),p_2(n),...,p_r(n)\rbrace$ where $0 \leq p_i(n) \leq 1$, $\Sigma p_i(n)=1$ are the action probabilities used for this choice. Based on the response of the enviornment, the probabilities $p_i(n)$ are updated for all i. Depending upon the qualitative objectives sought after, the desired asymptotic behavior of $p_i(n)$ (for all i) can be specified. In particular, if the objective is to converge to the "best" action, the "learning algorithm" has to be such that the probability $p_{opt}(n)$ corresponding to the best action converges to 1 in some stochastic sense, while $p_{i}(n)$ (where $d_i \neq d_{opt}$) converges to zero.

\underline{More General Stochastic Automata} can be described by a sextuple of the form ($\boldsymbol{\beta}$, $\boldsymbol{\Phi}$, $\boldsymbol{\alpha}$, $\boldsymbol{P}$, $\boldsymbol{A}$, $\boldsymbol{G}$) where $\beta$ is the input set, $\boldsymbol{\Phi}=\lbrace\phi_1,\phi_2,...,\phi_s\rbrace$ is the set of internal states $\boldsymbol{\alpha}=\lbrace\alpha_1,\alpha_2,...,\alpha_r\rbrace$ is the output or action set, q is the state probability vector $(q_1(n),...,q_s(n))$  corresponding to the 's' states, $\boldsymbol{A}$ is the learning algorithm which generates q(n+1) from q(n) and the response, and $\boldsymbol{G}$: $\boldsymbol{\phi} \rightarrow \boldsymbol{\alpha}$ is the output function. The above representation is merely given for the sake of completeness, but almost all the important results in the literature [2], have been derived using directly the action probability vector p(n). In this paper, we will consequently confine ourselves to the latter.

\underline{Psychology and Engineering:}
In the context of psychology, a learning automaton may be regarded as a model of the learning behavior of an organism. In such a case, the environment is controlled by the experimenter. In engineering applications such as pattern recognition, identification, or control (more generally machine learning), the controller corresponds to the automaton, while the rest of the system, with all uncertainties, constitutes the environment.

\underline{P,Q, and S Models} So far, we have assumed that the output of the environment (or the input to the automaton) is a binary set (0,1) i.e a penalty or a reward. We shall refer to such a model as a P-model. If the input is a finite set (say $q_1,q_2,...q_m$) with $0 \leq q_i \leq 1$, we shall refer to it as a Q model. If, however, the input to the automaton (or the output of the environment) can be any continuous function, we shall refer to it as an S model.

\underline{Comment:} While it may be desirable to use Q models or S models in practical applications, the principal concepts can be explained more easily and succinctly using P-models.

\underline{Norms of Behavior}
While designing learning schemes, an important question that arises even at the initial stages is whether the updating is done in such a manner as to result in a performance compatible with intuitive notions of learning.

Initially, when learning starts, it is natural to assume that all actions are chosen with the same probability i.e ($p_i(0)=1/r$). This implies that the average reward is $M_0=1/r\sum\limits_{i=1} d_i$

Hence, if the probabilities $p_i(n)$ are varied on-line, the question arises as to whether M(n), the average reward, is increasing. If $\lim_{n \to \infty} M(n) > M_0$, the learning scheme is said to be expedient. If $\lim_{n \to \infty} E[M(n)]=d_{opt}$, the automaton is said to be optimal. In practice, obtaining strictly optimal schemes is a very difficult undertaking.

\underline{Comment:} If the probabilities of all the actions other than that of the optimal tend to zero, the automaton will never choose them asymptotically. However, comparison with such responses is needed to assure that the action chosen is indeed optimal.
The above considerations lead to the definition of $\epsilon$- optimality.

\underline{Definition} A learning scheme is said to be $\epsilon$-optimal if
\begin{equation}
\lim_{n \to \infty} E[M(n)] =d_{opt}-\epsilon
\end{equation}
can be achieved for any arbitrary $\epsilon>0$, by a suitable choice of the parameters of the reinforcement schemes. $\epsilon$-optimality consequently implies that the performance of the learning automaton can be made to be as close to the optimal as desired (but not zero).

Finally, if E[M(n)] increases monotonically i.e. 

\begin{equation}
E[M(n+1)|p(n)]>M(n)
\end{equation}

for all n and all $d_i$, the automaton is said to absolutely expedient.
\section{Review of Learning Automata Schemes}
In the previous section we defined a stochastic learning automaton and defined some norms by which the performance of the different learning schemes can be evaluated. The different schemes represent different sequential choices of actions out of an input set, to improve the responses from a random environment. In this section we present, very briefly, the most significant schemes out of the set of all schemes that have been proposed in the literature for this problem. Perhaps more important are the mathematical results related to the convergence of the different schemes, since we will be concerned with similar issues while dealing with the principal topic of this paper i.e. the effect of the use of multiple models on the speed of  convergence of learning schemes. For detailed treatment of all aspects of learning automata schemes, the reader is referred to the book by Narendra and Thathachar [2].

\subsection{Reinforcement Schemes}
In general, most of the reinforcement schemes that have been proposed in the past can be represented by the difference equation
\begin{equation}
p(n+1) = T(p(n),\alpha(n), \beta(n))
\end{equation}
where T is a mapping and $\alpha (n)$ and $\beta (n)$ are respectively the input chosen and the response obtained form the environment at time 'n'. T determines how this information is to be used for choosing p(n+1) at time (n+1).

\underline{Linear and Nonlinear Schemes:} If T is linear, the scheme is referred to as a linear scheme. Similarly, nonlinear and hybrid schemes can also be defined (in the latter case two or more schemes are combined).
 
\underline{Asymptotic Behavior:} Learning Schemes can also be classified on the basis of their asymptotic behavior as expedient, $\epsilon$-optimal or optimal.

\underline{Ergodic and Non-ergodic:} The theory of Markov processes forms the principal vehicle for the study of learning schemes. When these schemes are used in stationary environments, they result in Markov processes that are either ergodic or contain absorbing states. Hence ergodic and non-ergodic schemes are also terms used to describe learning automata schemes.

\subsection{General Reinforcement Schemes:}
General reinforcement schemes for updating probabilities can be represented as

$\alpha(n)=\alpha_i\;\;\;\;\;(i=1,2,...r)$

$p_j(n+1)=p_j(n)-g_j[p(n)]$ when $\beta(n)=1$

$\;\;\;\;\;\;\;\;\;\;\;\;\;\;\;\;=p_j(n)+h_j[p(n)]$ when $\beta(n)=0\;\;\;\;\;\;\;\;\;\;\;\; (5)$

$j \neq i$

We note that when action $\alpha_i$ is performed at instant 'n', the output $\beta(n)$ can be either a reward $\beta(n)=1$ or a penalty $\beta(n)=0$.
Qualitatively, this would suggest that the probability  $p_i(n)$ be increased in the former case and decreased in the latter case. However, it is seen in equation (5) that this is mathematically represented by a decrease or increase in the probabilities of the actions not chosen. With a reward, the probabilities of (n-1) actions are decreased, and that determines the increase in the probability of the action chosen. To conserve probability measure $p_i(n+1)=p_i(n)+\sum_{j \neq i} g_j (p(n))$. Similarly for a penalty $p_i(n+1)=p_i(n)-\sum_{j \neq i} h_j (p(n)) $. $g_j$ and $h_j$ are continuous non-negative functions and satisfy the inequalities $0<g_j(p)<p_j$ and $0<\sum_{j=1, j \neq i}^r[p_j+h_j(p)]<1$.

\underline{Comment:} A very large number of learning schemes have been proposed by different authors. If $g_j$ and $h_j$ are linear functions, the learning schemes are linear. If even one of the 2n functions is nonlinear, the scheme is defined as nonlinear. In this brief introduction to learning schemes, we shall consider only linear schemes. which are $\epsilon$-optimal or ergodic.

\underline{Linear Reward-Inaction ($L_{R-I}$) scheme}
If the automaton has two actions, the probabilities $p_1(n)$ and $p_2(n)$ are modified only when there is a reward. For example if $\alpha(n)=\alpha_1$ results in a reward, the probabilities are updated as:

$p_1(n+1)=p_1(n)+a(1-p_1(n))=a+(1-a)p_1(n)$

$p_2(n+1)=(1-a)p_2(n)=(1-a)(1-p_1(n))$

For a penalty output, no changes are made in all the probabilities. (i.e. penalty responses are ignored)

If the learning scheme has 'r' actions and $\alpha(n)=\alpha_i$ results in a reward
$p_j(n+1)=(1-a)p_j(n)$

$p_i(n+1)=1-(1-a)\sum_{j \neq i}p_j(n)$ and $p_j(n)=p_i(n)+a(1-p_i(n))$

\underline{Comment:} The $L_{R-I}$ scheme is an $\epsilon$-optimal scheme and has absorbing states (two in a two action scheme and r unit vectors in an r-action scheme).

\underline{$L_{R-P}$ Scheme:} If the action probabilities are increased for a reward and decreased for a penalty, we have ergodic schemes which do not have absorbing states. $L_{R - \epsilon P}$ schemes were proposed to have many of the advantages of $L_{R-I}$ schemes, even while enjoying the property of ergodicity. In this case reward and penalty are not treated symmetrically, with changes (increase or decrease in probability) small for a penalty output, compared to the changes when the output is a reward. As is to be expected, the convergence properties of ergodic schemes are very different from those of the $L_{R-I}$ scheme.

Since our principal interest in this and the following papers is in the use of multiple models for improving convergence rates of learning algorithms, we now consider simulation results obtained using $L_{R-I}$, $L_{R-P}$, and $L_{R-\epsilon P}$ schemes.
\section{Simulation Results}
Numerous linear and nonlinear learning algorithms have been proposed in the literature by various authors, and extensive simulation studies have been carried out on the computer on all of them. Our objective in this section is not to discuss in detail all the proposed schemes, but merely to provide the reader with an understanding of the factors that govern the speed and accuracy of some of the more commonly used schemes. The simulations included in this section will serve as benchmark examples to be used for comparison purposes while discussing alternate learning schemes.

\begin{figure}[H]
\includegraphics[width=0.5\textwidth,height=0.25\textwidth]{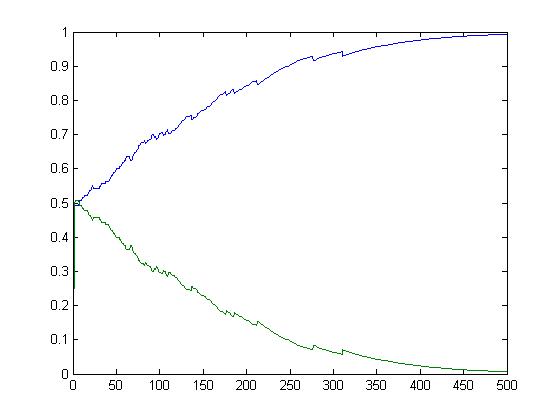}
\caption{$L_{R-I}$ with 2 actions for a= 0.015}
\label{Fig2}
\end{figure}

\begin{figure}[H]
\includegraphics[width=0.5\textwidth,height=0.25\textwidth]{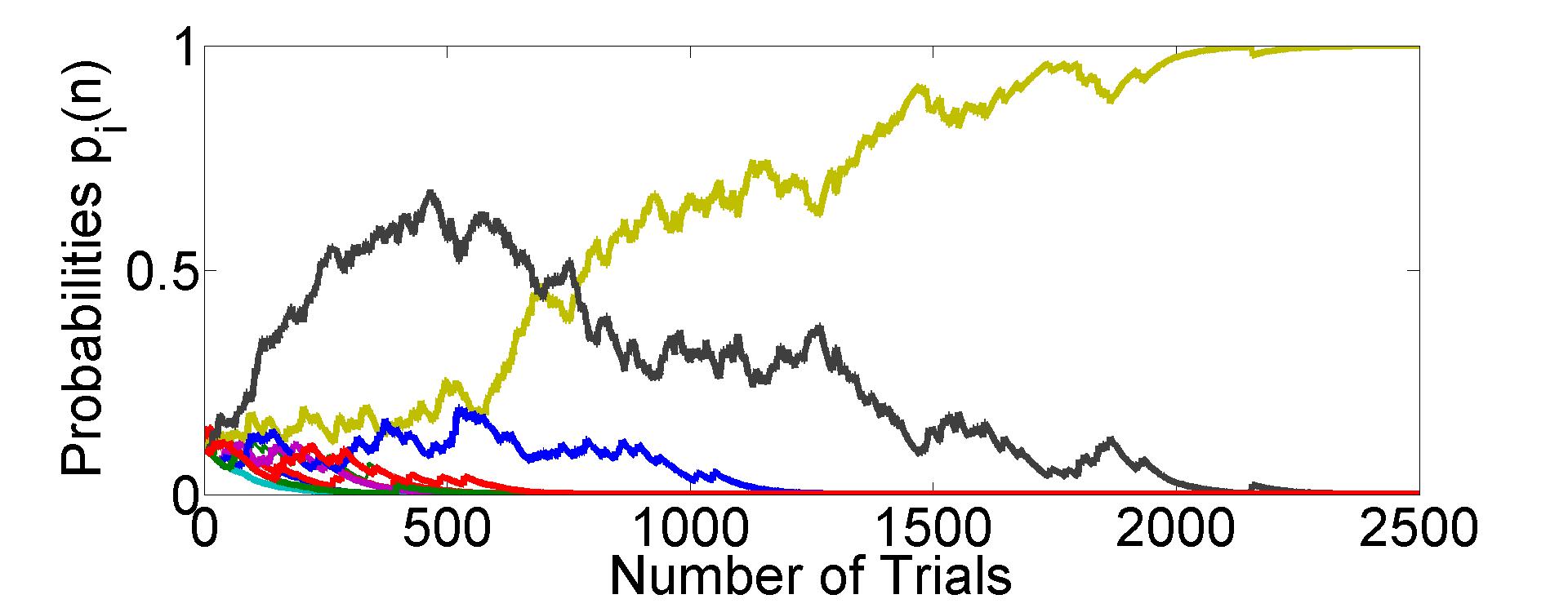}
\caption{$L_{R-I}$ with 10 actions for a= 0.015}
\label{Fig3}
\end{figure}

\begin{figure}[H]
\includegraphics[width=0.5\textwidth,height=0.25\textwidth]{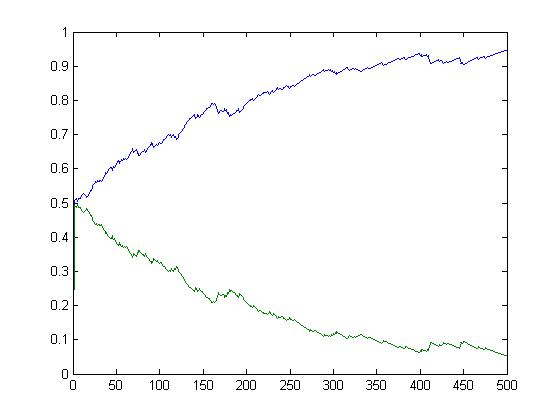}
\caption{$L_R-P$ with 2 actions for a= 0.015, b=0.005}
\label{Fig4}
\end{figure}

In Figure 2, the simulation study of an $L_{R-I}$ algorithm with two actions is shown. The only parameter that can be adjusted in this case is "a". For the experiment, a=0.015 was chosen. The convergence time for a typical simulation is seen to be approximately  500 steps.

Increasing "a" improves the rate of convergence, but also increases the probability of convergence to the wrong action.

In   Figure 3, a ten-action case is considered and the convergence time on a sample path is seen to be 2000 steps. As stated earlier, $L_{R-P}$ schemes are ergodic and hence sample paths do no converge to any fixed probability as in the $L_{R-I}$ case, but converge in distribution. Once again, for a two-action case , the probability of the best action reaches 0.95 in 500 steps.

\underline{Comment:} In the following sections, we will be interested in the improvement in convergence that can be achieved using modifications in the learning schemes (i.e 500 steps for 2 actions and 2000 steps for 10 actions).
\section{Other Learning Schemes}
Thus far we have discussed (direct) learning algorithms in which at the end of every trial the probability vector p(n) is updated based on the response of the environment. There are also a number of other updating schemes proposed in the literature for variable structure stochastic automata. In the  "Special Issue on Learning Automata of the Journal of Cybernetics and Information Science" Edited by the first author [3], Tyspkin and Poznyak (1977) attempt to unify the various learning algorithms within the general framework of stochastic approximation. Thathachar and Sastry (1985) [4] incorporate estimates of the reward probabilities in the updating schemes and prove $\epsilon$-optimality. Such schemes have been called estimator algorithms and have a higher rate of convergence in stationary random environments as compared to $L_{R-I}$ and $L_{R-P}$ schemes. One version of the estimator algorithm is called the pursuit algorithm and is described in the following section.
\subsection{The Pursuit Algorithm}
In the $L_{R-I}$ or $L_{R-P}$ schemes, the efficiency of an action was judged on the basis of the output produced by the action at one instant of time. However, our interest is in the action $\alpha_{opt}$ with the largest reward probability $d_{opt}$. To estimate this, the effect of every action over all the past attempts is stored in this procedure. If $\bar{d}_i$ is the estimate of  $d_i$ based on all the past attempts of the $i_{th}$ action $\alpha_i$, the probability vector p(n) is adjusted at every instant in the direction of the current optimal action based on these estimates. The three steps in the procedure are listed below:\\
(1) based on the probability distribution p(n) at instant 'n' an action $\alpha (n)$=$\alpha_i$ is chosen which produces a reward or a penalty.\\
(2) the estimate $\bar{d}_i(n-1)$ of the reward probability is updated on the basis of the above response to $\bar{d}_i(n)$. If the highest value of $\bar{d}_i(n)$ is $d_j(n)$(j=1,2,...,r), the action probability vector p(n) is modified as
\begin{equation}
p(n+1)=p(n)(1-\lambda)+\lambda l
\end{equation}
where $\lambda$ is a scalar with $0<\lambda<1$, and l is a unit vector with unity as the $i_{th}$ element and all other elements zero. This orients the vector p(n) more towards the optimal action.

\underline{Comment:} In the 1960s, in adaptive control, it was realized that adaptive parameters should be adjusted on the basis of all the past data, rather than instantaneous values. The modification described here for learning algorithms was motivated by this.

The pursuit algorithm can also be shown to be $\epsilon$-optimal, and is significantly faster than the $L_{R-I}$ and $L_{R-P}$ schemes as shown in the simulations included below for two action and ten action schemes. 

\begin{figure}[H]
\includegraphics[width=0.5\textwidth,height=0.25\textwidth]{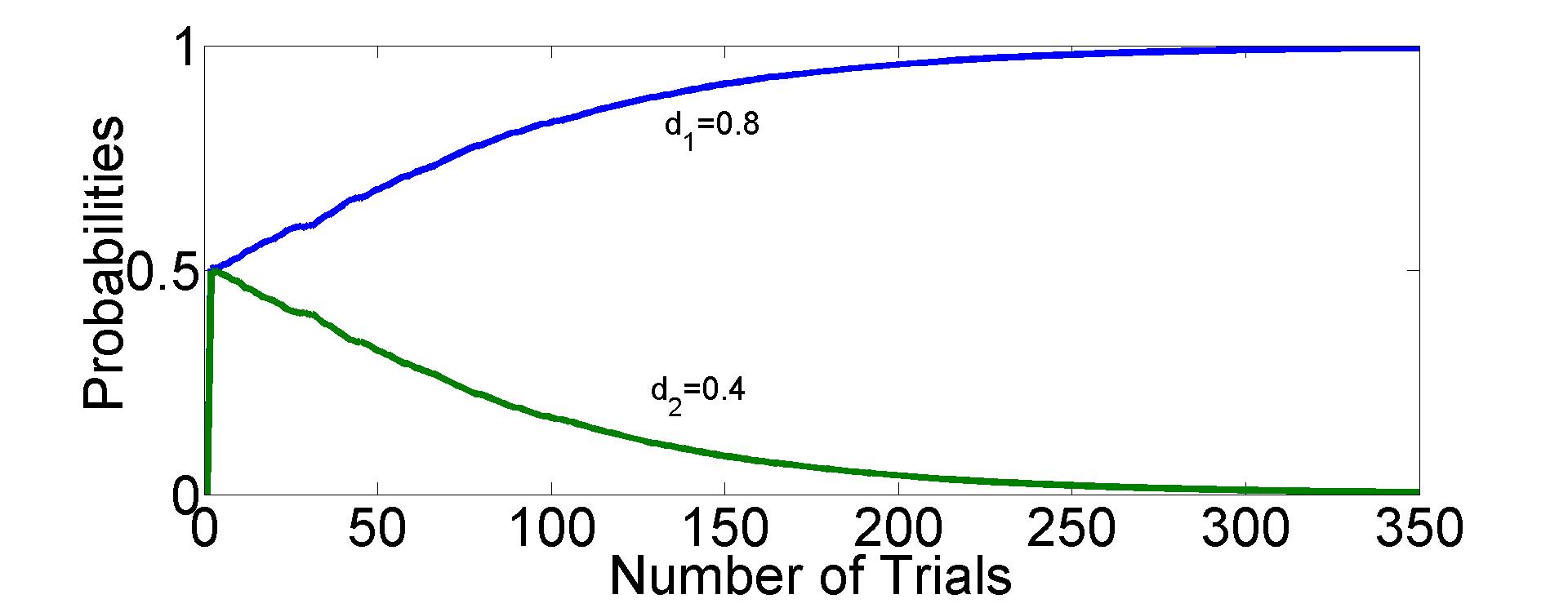}
\caption{Pursuit algorithm with two actions}
\label{Fig5}
\end{figure}

\begin{figure}[H]
\includegraphics[width=0.5\textwidth,height=0.25\textwidth]{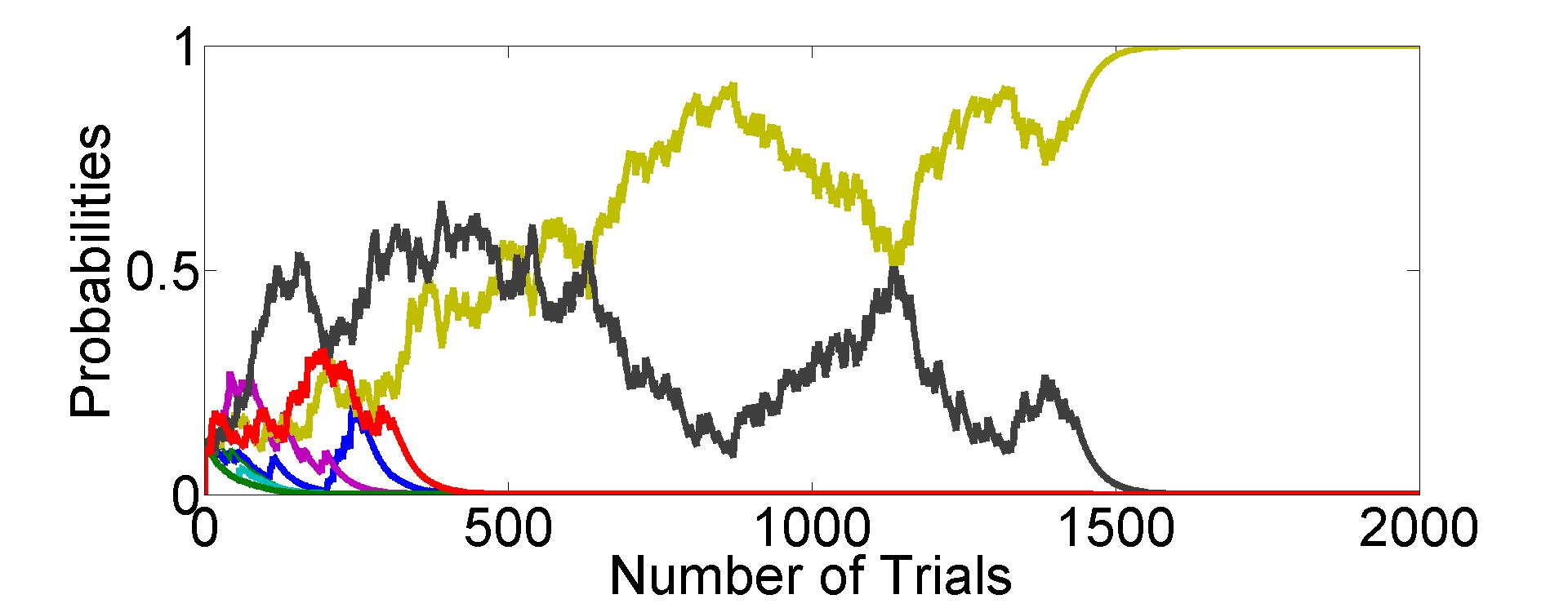}
\caption{Pursuit algorithm with ten actions}
\label{Fig6}
\end{figure}

Figure 5 and 6 indicate typical responses of a two action and a ten action learning automaton using a pursuit algorithm. It is seen that the typical times for convergence are 300 and 1500 respectively, which are significantly faster than obtained using those simple $L_{R-I}$ schemes.

\section{Learning Automata Using Multiple Models}
From the previous discussions, a model can be set up to estimate the reward probability of each action. Following this, a certainty equivalence principle can be used to decide which action appears to be the best, and the action probability vector can be moved by a small step towards the unit vector corresponding to that action.

\underline{Multiple Fixed Models}: 
Let $0<q_1<q_2<...<q_m<1$ be m fixed probabilities. Let these values represent fixed estimates of the reward $d_i$ corresponding to action $\alpha_i$. If similar models are also used for all the 'r' actions, there are a total of $mr$ probabilities that are used to determine the strategy at any instant.

Let the output of the $i^{th}$ action $\alpha_i$ at stage n consist of $n_1$ rewards and $n-n_1$ penalties. Then corresponding to any probability, (say $q_1$), the likelihood of the event is $q_1^{n_1}(1-q_1)^{n-n_1}=d_i(q_1)$. Similarly, $d_i(q_2), d_i(q_3), ..., d_i(q_m)$ are computed and the maximum $d_{i(opt)}(n)$ is determined. At this stage, since the maximum value of $d_{i(opt)}(n)$ for all i actions is known, the probability vector (as in the pursuit algorithm) is adjusted incrementally in the direction of the unit vector corresponding to that action.

The relation between the models described above and the pursuit algorithm has been studied extensively, but due to space limitations is not included here. Only the simulation for the 10 action case is shown in Figure 7. It is seen that there is a substantial improvement both in the speed of convergence as well as the smoothness of the optimal probability. The convergence time of 300 is seen to be a significant improvement over the 2000 steps needed with the $L_{R-I}$ algorithm, and 1500 steps needed with the pursuit algorithm.

The result given in Figure 7 are not conclusive but merely indicate that considerably more work needs to be done using multiple models.

\begin{figure}[H]
\includegraphics[width=0.5\textwidth,height=0.25\textwidth]{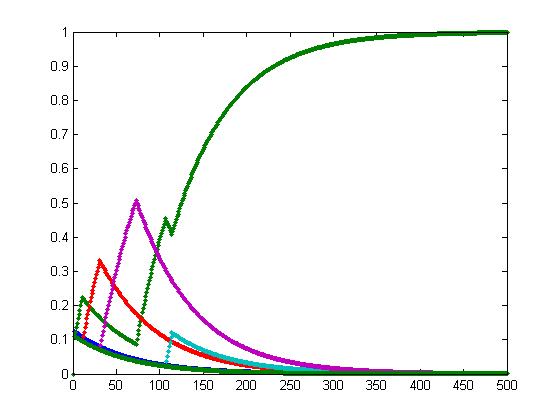}
\caption{Ten actions with Multiple Fixed Models}
\label{Fig7}
\end{figure}




\underline{Multiple Adaptive Models}:
Even though the learning automaton is a direct scheme, the procedure described above was used to estimate which of the fixed probabilities (models) was closest to the unknown reward probability corresponding to each action. A natural extension of the above procedure is to make all the models adaptive (i.e. adjust the probabilities $\lbrace q_1,q_2,...q_r\rbrace$) adaptively, on the basis of the responses of the environment to the various actions. It is immediately evident that all values must converge to the reward  $d_i$ for the $i^{th}$ action,  as 'n' tends to infinity. This fact can be made use of  to determine where $d_{opt}$ lies, and the action corresponding to it. This will be treated in detail in future papers.

%
%
%


\section{Multiple Model Reinforcement Learning in Dynamic Environments}

In adaptive control, multiple model based approaches have been very effective in indirect control i.e. those situations where identification of the process to be controlled precedes taking a control action. If a model of the process is known, it can be used to generate a control input which results in  the desired response of the model, and consequently (it is hoped) of the plant. If multiple models are used, methods for choosing the "best" model or " a combination of the models" have been discussed in the adaptive literature. Our objective is to use a similar approach in the future for learning schemes as well.

\underline{Comment:} In the previous section, "multiple models" were used for a direct learning scheme. This merely underscores the fact that the term "model" can be loosely interpreted as a description of the system which permits a decision to be made concerning its behavior.

In this section, we provide a sketch as to how multiple model approaches can be extended to reinforcement learning problems involving dynamic environments. In such a case, the details of the methods and their theoretical analysis are considerably more complex than those for a static environment involving learning automata, and will be described in detail in future papers.

\vspace*{0.3cm}
\underline{Problem Formulation:}
\vspace*{0.3cm}

\underline{Discrete-State}: A Markov Decision Process (MDP) is defined by the following quantities:
\begin{itemize}
\item 
	The State Space  $S$, a finite set,  
\item	The action set $A$, a finite set, listing all actions available to the agent in any state
\item	A state transition probability function $P : S  X S X A  \rightarrow  (0; 1)$
\item	An immediate payoff function  $R : S X S X A \rightarrow 0; 1]$ where 0 corresponds to no reward and 1 corresponds to reward.
\end{itemize}
The objective of the agent is to maximize the overall discounted reward, i.e., the objective is not merely to maximize the reward for the next transition, but to maximize the reward over all future transitions made from an initial state.

\vspace*{0.3cm}

\underline{Continuous-State}: A nonlinear dynamical environment is described by the equation
$$x(k+1) = f[x(k),u(k), n(k)]$$
where x(k), u(k), and n(k) are the state, agent action, and noise respectively at instant k.
The performance index is given by $$J=\sum_{(k=0)}^{\infty} \gamma^{k} R[x(k),u(k)]$$
where $0 < \gamma < 1$ is a discount factor.

The objective of the agent is to determine a policy $u(k)=g[x(k)]$ to optimize $J$.

In the model-based approaches to reinforcement learning, an identification model of the environment
(in the form of $\hat{P}$  and $\hat{R}$  matrices for discrete case, and $\hat{f}$  function in the continuous case) is estimated during on-line learning and this model is used in conjunction with dynamic programming or the Hamilton-Jacobi-Bellman equation to optimize J.

\underline{Multiple Model Approaches:} In the discrete-state MDP formulation, since each element of the $P$ matrix is a probability, multiple fixed or adaptive models can be set up to estimate them, and the best model can be selected in a manner similar to that used for a learning automaton. The best selected model(s) can then be used to carry out the dynamic programming computations. 

In the continuous state case,  instead of only one such model, a bank of identification models, as described below, is used:
$$\hat{x}_i (k+1)=\hat{f}_i [x(k),u(k),\theta_i (k)]     i=1, \ldots , N$$
and these models are combined in some manner to compute the predicted state, with the aim of achieving higher accuracy than each of the individual models. It is worth noting that the structure of the models need not necessarily be the same, but can in principle be heterogeneous. This approach, as stated earlier, has been a popular one in the adaptive control field, to improve transient performance of adaptive control systems. The authors believe that such an approach can also be used in reinforcement learning systems as one way to speed up learning. 

\section{Conclusions}

In this paper, a very simple learning scheme (i.e. the Learning Automaton) was introduced and the properties of different learning algorithms were discussed. It was then shown that approaches, similar to multiple model based approaches in adaptive control, lead to faster and smoother convergence than conventional algorithms. Some comments were made towards the end of the paper as to how the approaches may be extended to significantly more complex learning schemes in dynamic environments.


\begin{thebibliography}{1}

\bibitem{1}
Tsypkin, Yakov Zalmanovich. Adaptation and learning in automatic systems. Vol. 73. New York: Academic Press, 1971.
\bibitem{2}
Narendra, Kumpati S., and Mandayam AL Thathachar. Learning automata: an introduction. Courier Dover Publications, 2012.
\bibitem{3}
Special Issue on Learning Automata of the Journal of Cybernetics and Information Science, 1977
\bibitem{4}
L Thathachar, M., and P. Shanti Sastry. "A new approach to the design of reinforcement schemes for learning automata." Systems, Man and Cybernetics, IEEE Transactions on 1 (1985): 168-175.
\end{thebibliography}
\end{document}